# Real-World Single Image Super-Resolution Under Rainy Condition

Mohammad Shahab Uddin
*Department of Electrical and Computer Engineering*
*Old Dominion University*
Norfolk, USA
muddi003@odu.edu

*Abstract*— Image super-resolution is an important research area in computer vision that has a wide variety of applications including surveillance, medical imaging etc. Real-world signal image super-resolution has become very popular now-a-days due to its real-time application. There are still a lot of scopes to improve real-world single image super-resolution specially during challenging weather scenarios. In this paper, we have proposed a new algorithm to perform real-world single image super-resolution during rainy condition. Our proposed method can mitigate the influence of rainy conditions during image super-resolution. Our experiment results show that our proposed algorithm can perform image super-resolution decreasing the negative effects of the rain.

*Keywords—image super-resolution, rainy condition, deep learning*

## I. INTRODUCTION

Image super-resolution denotes the enhancement of the resolution of images from low resolution to high resolution. Image super-resolution is extremely important for improving performance of deep learning models during several situations including surveillance. Using super-resolution, we can improve the resolution of an image to get better classification and detection accuracy. Many methods are proposed for image super-resolution. But we have not found any work that has used real LR images during training to perform real-world single image super-resolution during extreme weather conditions like rain. The only work that has considered rain during image super-resolution is proposed in [1].But this work has used synthetic LR images for their method. So, we have decided to improve this area to get better performance of image super-resolution in rainy condition. Our method can perform image super-resolution during rainy conditions eliminating the effects of the rain on the image. Our model provides clean HR images from the rainy LR images.

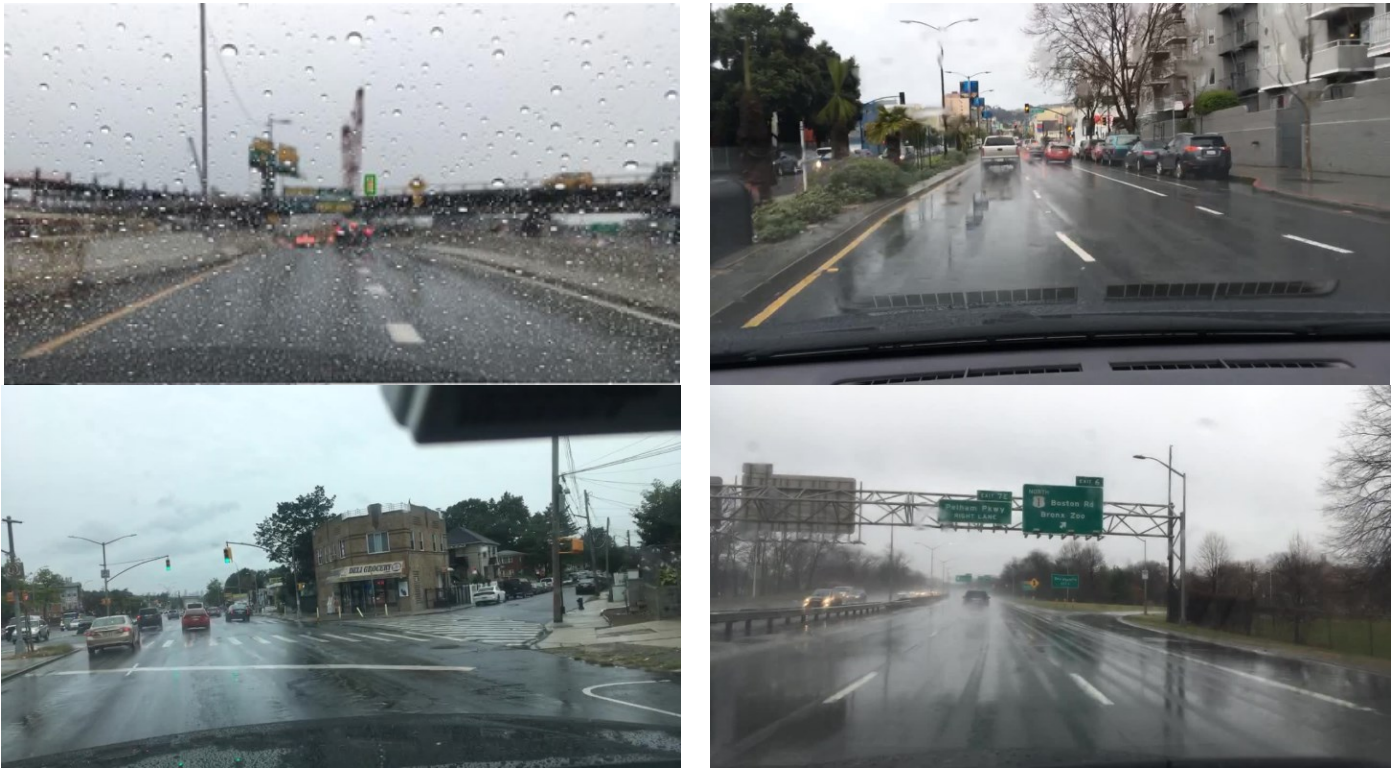

Fig. 1. Images under rainy condition.

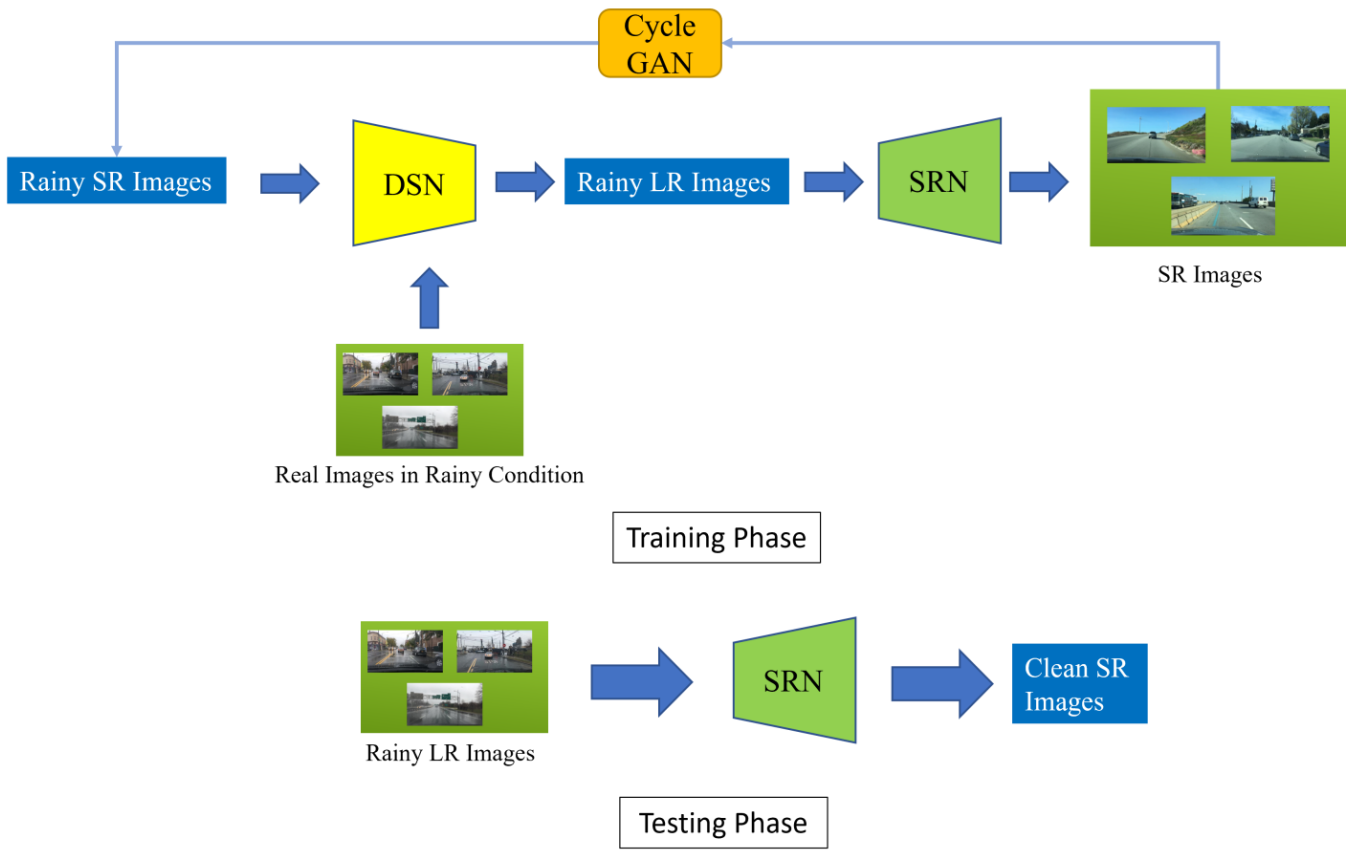

Fig. 2. Overview of our proposed method.

Rainy condition can create different kinds of negative effects on images. Rain can impair and distort the quality of an image. Rain consists of many drops which have different sizes and complex shapes. Moreover, raindrops cause intensity variation that can hampers the classification and detection accuracy. Also, rain streaks results in low contrast in visual data [2]. Figure 1 shows some images under rainy conditions. We can see the negative effects of rain on the images. Rain greatly effects the visual quality of the images that decrease the detection and classification accuracy of the machine learning model [2].

In our work, we have tried to mitigate the influence of rainy condition during image super-resolution. Our target is to get clean high-resolution image from the rainy low-resolution image. Our contribution is that we have proposed a new method that can perform real-world single image super-resolution during rainy condition.

## II. RELATED WORKS

Image super-resolution can be divided into two types [3] which are single image super resolution [4-6] and multi-frame image super-resolution [7-9]. In single image super-resolution, we only use one image to get high-resolution of that image. On the other hand, we have to use consecutive number of frames to get high-resolution of an image. Scientists have worked on various methods to perform signal image super-resolution including regression-based methods [10-11], reconstruction-based method [12-13], deep learning-based method [4-6]etc. [3]. Deep learning-based method has become very popular due to the increase of the performance of the deep learning models and computational devices. Real-world single image super-resolution emphasizes on the application during real-world situation. Researchers have proposed various methods for real-world single image super-resolution that include degradation modeling-based method, image pairs-based method, domain translation-based method, self-learning-based method etc. Degradation modeling-based methods [14-17] use synthetic low-resolution images for training a model for image super-resolution. These methods do not perform well when we use real LR images for super-resolution. For solving this issue, some works are done to collect paired LR-HR real images using camera [18-20]. Image pairs-based methods [18-20] require paired LR-SR images which are difficult to collect. Recently, some works have proposed generating real LR images using domain-translation methods [4-6,21-27] . These methods use generative adversarial networks [18] to generate LR images that are aligned with the domain of real LR images. For implementing this method, we do not need any paired dataset. Self-learning-based methods [28-30]are also not practical as these methods are time-consuming to train, and these methods do not use any external information to train model.

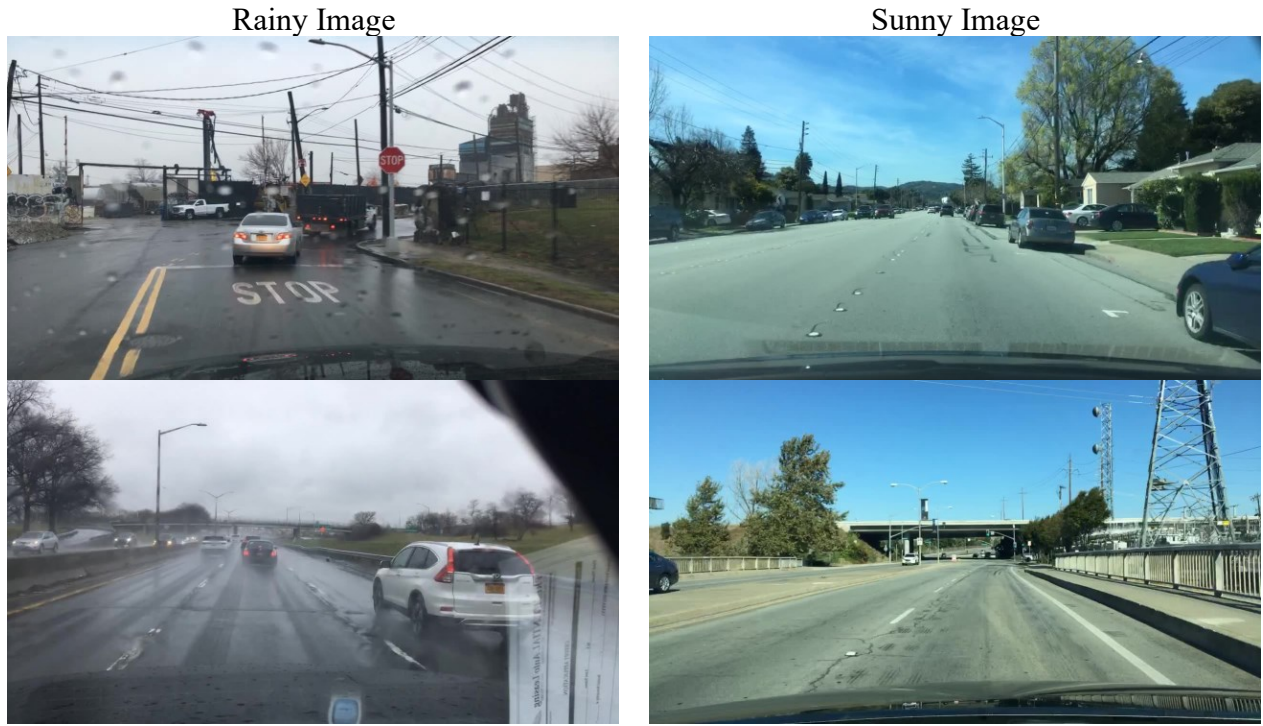

Fig. 3. Collected data from BDD100K dataset.

Overall, domain translation-based method is a good candidate for real-world single image super-resolution as we can use unpaired LR-SR images for training the model. As, these methods incorporate domain information of real LR images during training, so these methods can be used for real-time application. We have found a drawback of the image super-resolution methods is that they have not considered extreme weather situation like rain for real-world single image super-resolution during real-time.

Image super-resolution under extreme condition was proposed in [1] , but they have used synthetic LR images for training their deep learning model. They did not consider real LR images during training their model. So, this model is not perfect for using in real-time situation.

In [4], researchers have used GAN to generate LR images from SR images at first. After that, they used paired LR-SR images to train SR model. They have not considered rainy condition during the training. So, when we will perform image super-resolution using their method the HR image will have the negative effects of the image.

Recently, [5] has proposed a model to perform image super resolution using domain translation-based method. They used down sampling network and super-resolution network to train the model. They have used domain distance map to get better results. But they also have not considered extreme conditions like rain. So, the model will also provide a HR image that will contain the effects of the rain.

In [26], authors have used a multiple Cycle-in-Cycle network structure to perform image super-resolution using unpaired dataset. They have used multiple adversarial networks in their model. The main drawback of their method is that they have also not considered extreme conditions like rain. They have not worked to remove rainy effect during image super-resolution.

In [25], an unsupervised method was proposed for image super-resolution using unpaired dataset. They used CycleGAN and domain discriminator in their model. They also included various losses in their algorithm including pixel-wise loss, VGG feature loss. They also have not considered rainy condition for their model. This is the drawback of their method.

There are also many Deraining algorithms that are proposed to remove rainy effects from an image. As far we know, there is not any Deraining algorithms that have considered Deraining and Image super-resolution together. In [31], authors have used synthesis images for training the model. After that test the model on the synthesis images as well as real images. During training they also used a rain-streak binary Map for training the model to remove rain. On the other hand, our method does not need any paired dataset. We have trained our model with unpaired dataset to remove rain streak as well as perform image super-resolution

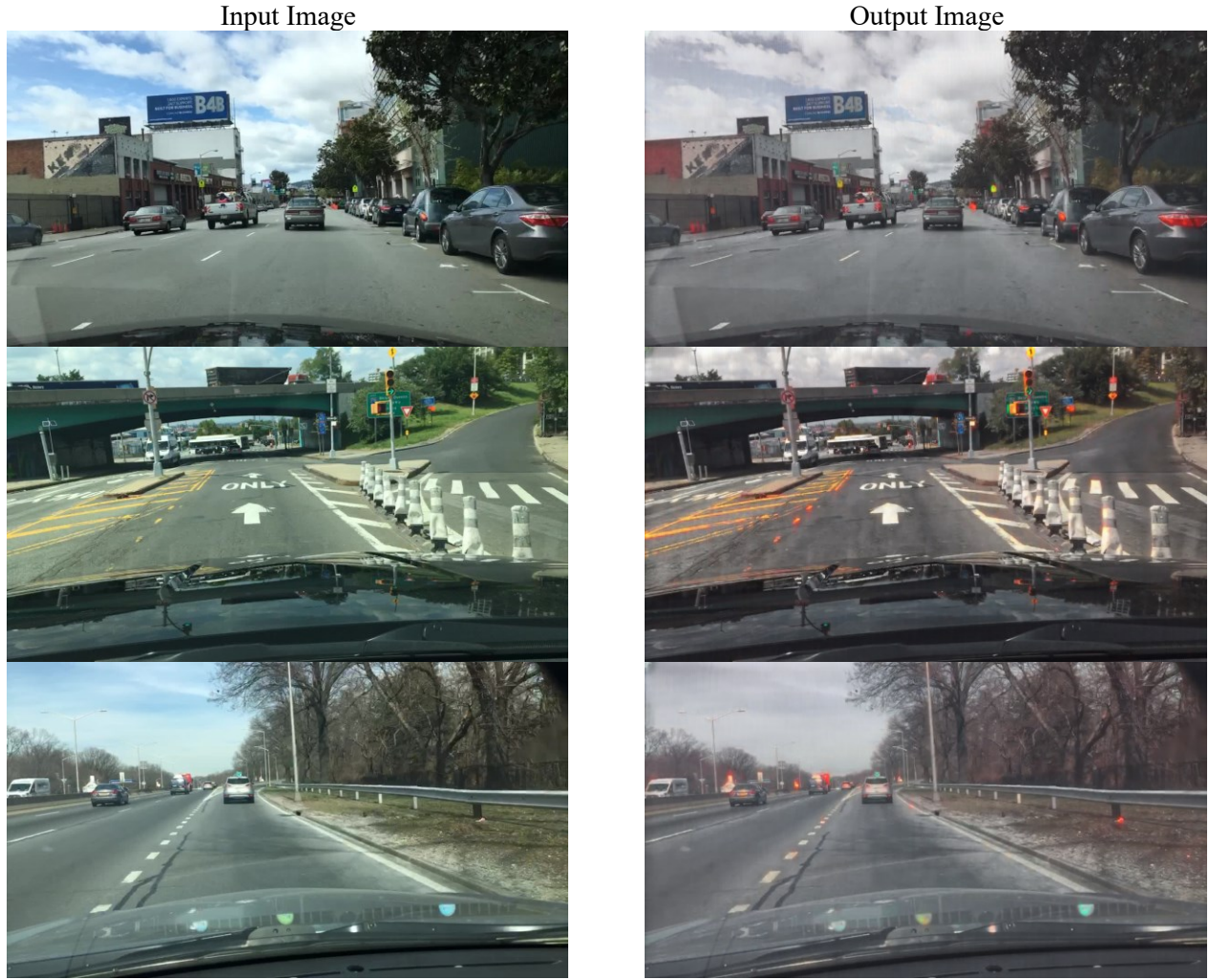

Fig. 4. Converting Sunny images to rainy image using CycleGAN

In [32], authors have proposed an attentive generative network for removing rain from a single image. They inject attention into generator and discriminator to remove the rain from the image perfectly keeping background information intact. But, main drawback of the model is that they need paired dataset to train the model. Our method does not need any paired dataset. Our method works with unpaired dataset for Deraining as well as image super-resolution together.

In [33], authors have used paired dataset for training a deep learning model for deraining an image. They proposed a model that learns about rain content at different scales and uses that information to get clean image. They also did not consider image super-resolution during training. Also, the main drawback of their method is that they need paired dataset.

In [34], authors have proposed a memory-oriented semisupervised (MOSS) method for deraining an image. Their model explore the characteristics of rain streaks form real and synthetic images. Their methods achieved better performance than the previous works but they have not performed image super-resolution during deraining.

In [35], authors have proposed an unpaired learning based method for rain remover from the image. Their proposed method consists of a knowledge distillation part and a semi-supervised learning part. They explored the properties of rain maps during training the model. Though their method has worked well during deraining, they have not considered image super-resolution during deraining.

Based on the review of these works, we have found there is not enough work that have performed image super-resolution of real LR images and at the same time mitigated the influence of the rainy condition. In our work we have solved this issue. We have performed image super-resolution and deraining together for a rainy LR image.

III. PROPOSED METHOD

*A. Proposed Model*

Our proposed algorithm is based on domain-translation based method. Our method involved three main steps during training: (a) generating the rainy images from sunny images using GAN, (b) generating LR rainy images using down sampling network, (c) generating SR images from LR images

LR HR (4×)

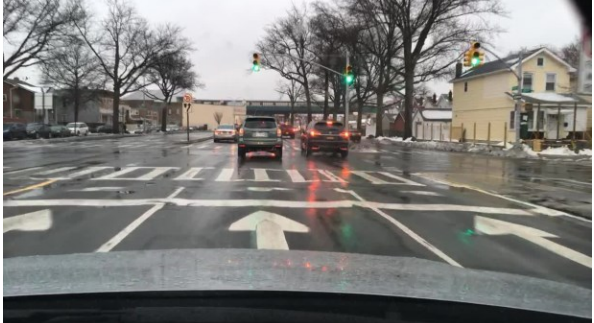 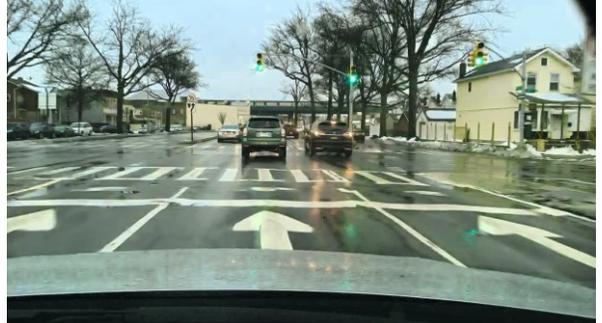
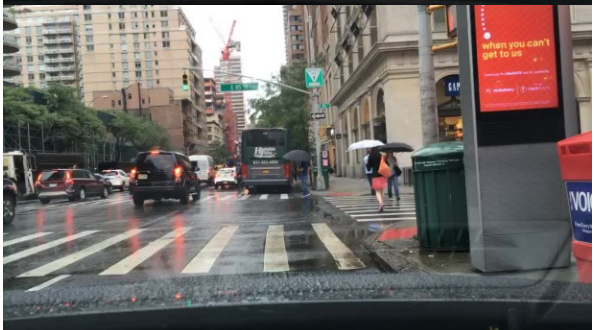 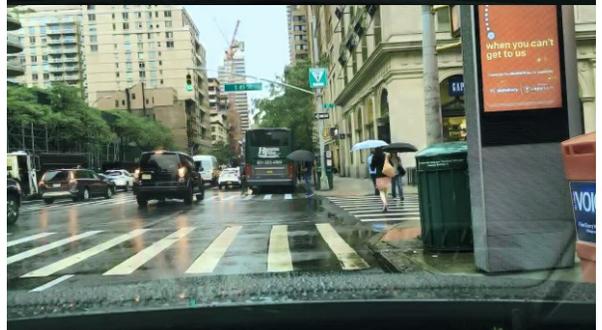
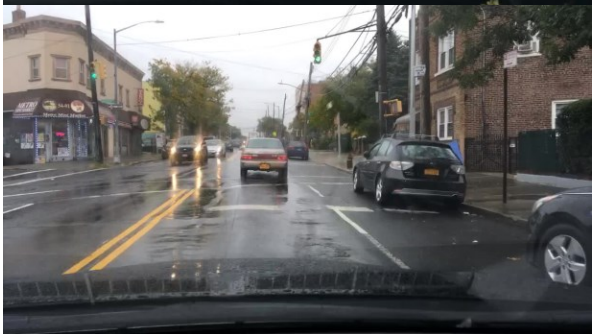 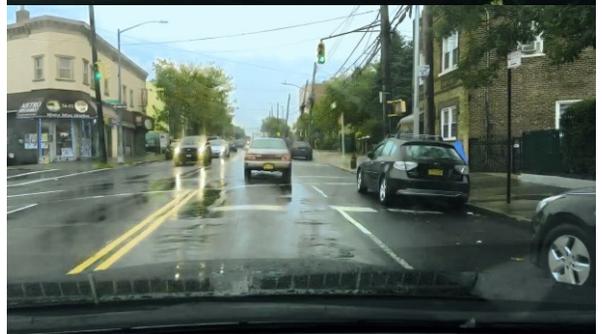
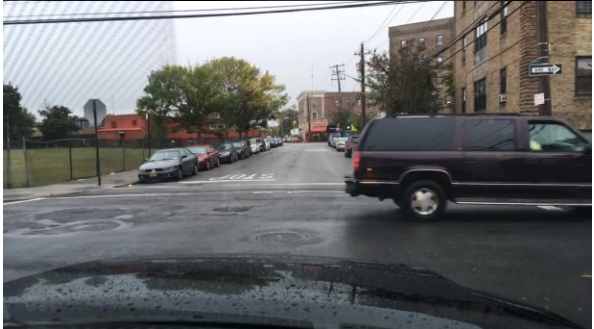 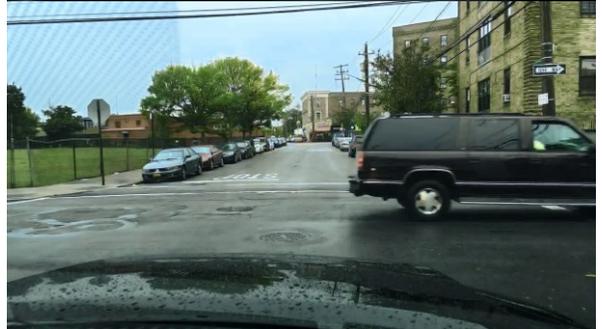
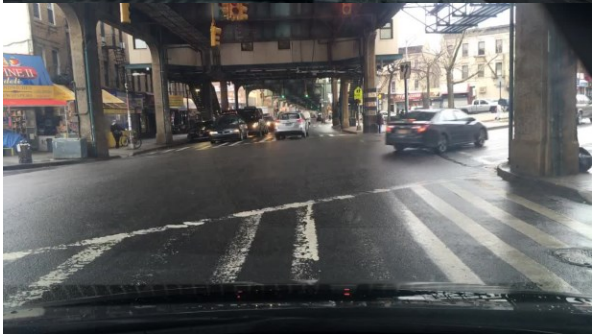 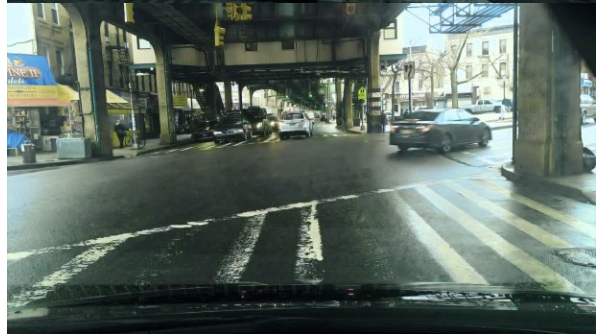

Fig. 5. Converting Rainy LR images to Clean HR image (4×) using SRN

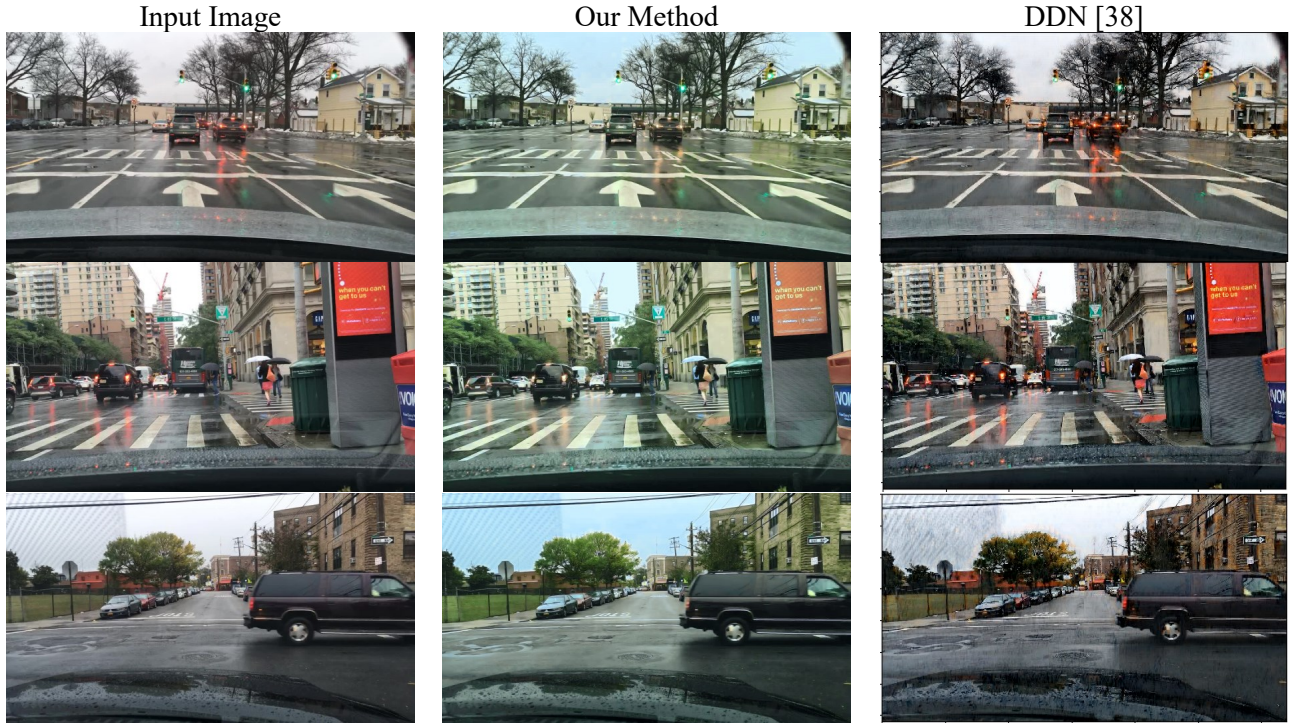

Fig. 6. Comparison between our method and DDN [38] with respect to the Deraining ability.

using super resolution network (SRN). We have performed conversion of sunny images into rainy images using Cycle GAN [36]. After that we have used DSN network proposed in [5] to generate LR rainy images from the generated rainy images. We reduced the size of image by 4 to get LR rainy image. Finally, we have used super resolution network to generate HR images. For SRN, we have used architecture mentioned in [5]. During training DSN and SRN, we have followed the steps mentioned in [5]. Figure 2 shows the overview of our proposed method.

### B. Data Collection for Training

We have used BDD100K dataset [37] for our experiments. This is a diverse driving dataset that was proposed for multitask learning. This dataset has images under various weather condition including sunny, rainy, snowy etc. We have collected 306 rainy images and 344 sunny images from the dataset. Figure. 3 shows some samples of the collected dataset.

### C. Training Details

We have trained the Cycle GAN for 3750 epochs using unpaired sunny and rainy images. DSN network is trained with converted rainy images from Cycle GAN and real LR images. We reduced the generated rainy images by 4 to get LR rainy image. Then, SRN is trained with generated LR rainy images from DSN and real HR sunny images. We have trained our model such that using SRN we can increase resolution of an image four times.

## IV. EXPERIMENTAL RESULTS

Figure 4 shows the output results from the cycle GAN during training. We can see that cycle GAN has able to convert the sunny clean images into the rainy images. We used these generated rainy images during training. After training the whole model, our SRN network is able to perform mage super-resolution on the LR rainy image directly. Figure 5 shows the Figure 5 shows the output results after performing image super resolution on LR rainy images. We notice that SRN has able to perform image super-resolution by increasing the resolution of the images by 4 times and at the same it was able to remove the negative effects of the rain from the image. We can see that the sky was cloudy in the original image, but the sky is clear in the output image. We have also compared our results with a deraining method called DDN [38]. DDN does not perform image super-resolution. So, we have compared the results with respect to the deraining ability. DDN needs paired dataset to train the model. They trained their model with paired dataset, and tested the model on the real rainy images. So, we have used their pre-trained model to test using our images. In figure 6, we can see that DDN has not performed well during deraining process. On the other hand, our method has performed better than the DDN during deraining. Also, our model performed image super-resolution.

## V. CONCLUSION

In this paper, we have proposed a new algorithm to perform image super-resolution under rainy conditions. Our proposed algorithm can perform image super-resolution and at the same time it can eliminate the negative effects of the rain from the image. In future, we have plans to extend this work to perform image super-resolution under other extreme weather conditions.

## ACKNOWLEDGMENT

We would like to thank Dr. Chen for his valuable feedbacks on this work.